\documentclass[10pt,journal,compsoc]{IEEEtran}

\usepackage{amsmath,amssymb,amsfonts}
\usepackage{graphicx}
\usepackage{booktabs}
\usepackage{multirow}
\usepackage{array}
\usepackage{cite}
\usepackage{url}
\usepackage{xcolor}
\usepackage{tikz}
\usetikzlibrary{arrows.meta,positioning,fit,calc}
\usepackage[caption=false,font=footnotesize]{subfig}

\newcommand{\R}{\mathbb{R}}
\newcommand{\C}{\mathbb{C}}
\newcommand{\Tr}{\operatorname{Tr}}
\newcommand{\method}{LegoQ}
\newcommand{\rhoavg}{\bar{\rho}}
\newcommand{\proj}{\mathcal{P}}

\begin{document}

\title{LegoQ: Density-Matrix Representation Learning with Spectral-Spatial State Transitions for Hyperspectral Image Classification}

\author{Weijia~Cao, Xiaofei~Yang, Fu~Wang, Yicong~Zhou, and Xiang~Zhou}

\markboth{Preprint}{Cao \MakeLowercase{\textit{et al.}}: LegoQ for Hyperspectral Image Classification}

\maketitle

\begin{abstract}
Hyperspectral image classification is complicated by mixed pixels, spectral ambiguity, class imbalance, and limited annotations. Most current classifiers encode a pixel or patch as a deterministic vector and apply a linear or multilayer softmax head. Although effective for discrimination, this representation does not directly expose how mixed or uncertain a sample is. This paper presents \method, a classical density-matrix representation learning framework for hyperspectral images. The spectral bands are divided into groups and each group is mapped to a positive semi-definite, Hermitian, trace-normalized matrix state. A composable stack of spectral, spatial, and inter-group transitions then updates the states while repeatedly projecting them back to the valid state set. Instead of flattening the final features, \method\ aggregates the group states and compares them with learnable class-prototype density matrices through Uhlmann fidelity. The normalized eigenspectrum, von Neumann entropy, purity, and prototype fidelity provide sample-level diagnostics that are unavailable from a conventional vector head. On Indian Pines, ten runs yield an overall accuracy of $96.20\pm0.70\%$, an average accuracy of $95.57\pm1.29\%$, and a kappa coefficient of $95.66\pm0.80\%$. On WHU-Hi-LongKou, the best of ten runs reaches $97.52\%$ overall accuracy. Classification maps and feature projections show that the transition stack produces compact and better separated class structures. The results support constrained matrix-state learning as a practical alternative to vector-only hyperspectral classification without requiring quantum hardware.
\end{abstract}


\section{Introduction}
\IEEEPARstart{H}{yperspectral} images record a densely sampled spectrum at each spatial location. The resulting data cube supports fine material discrimination in agriculture, environmental monitoring, and urban analysis, but it also creates difficult recognition conditions. A single pixel may contain multiple materials, adjacent classes may exhibit similar spectra, and labeled samples are commonly sparse and imbalanced~\cite{landgrebe2003signal,fauvel2013advances,bioucas2012unmixing}. These effects are pronounced around object boundaries and heterogeneous regions, where a point representation can hide the mixture structure that generated the observation.

Deep hyperspectral image (HSI) classifiers have progressed from spectral convolution to joint spectral-spatial convolution, residual learning, transformers, and selective state-space models~\cite{chen2014deep,chen2016cnn,zhong2018ssrn,roy2020hybridsn,hong2022spectralformer,gu2023mamba,li2024mambahsi}. Their feature extractors differ substantially, but their last stage is usually similar: a tensor is flattened or pooled into a vector, followed by a linear or multilayer softmax classifier. Such classifiers provide a decision score, but the score is not itself a measurement of representation mixedness. In particular, a low softmax confidence may arise from class overlap, limited training, poor calibration, or mixed materials, and these causes cannot be separated from the vector output alone.

We study a different representation space. A density matrix is Hermitian, positive semi-definite (PSD), and normalized to unit trace~\cite{nielsen2010quantum}. The trace constraint converts its eigenvalues into a normalized spectrum. This makes entropy and purity comparable across samples and gives the off-diagonal entries a direct role as cross-mode correlations. We use this mathematical structure as a classical neural representation; the method does not require quantum hardware and does not claim quantum computational advantage.

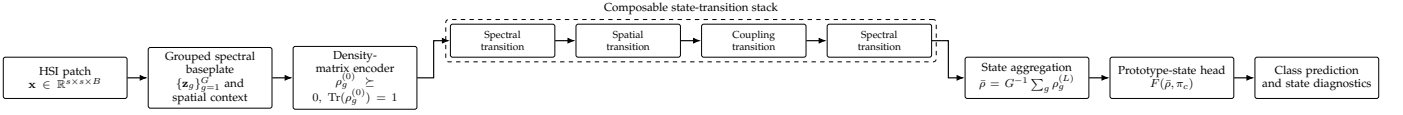
\begin{figure*}[t]
\centering
\resizebox{\textwidth}{!}{%
\begin{tikzpicture}[
    node distance=4mm and 5mm,
    box/.style={draw, rounded corners=1.5pt, align=center, minimum height=10mm, text width=28mm, font=\footnotesize, inner sep=3pt},
    smallbox/.style={draw, rounded corners=1.5pt, align=center, minimum height=8mm, text width=24mm, font=\scriptsize, inner sep=2pt},
    arrow/.style={-{Latex[length=2mm]}, thick},
    group/.style={draw, dashed, rounded corners=2pt, inner sep=3pt}
]
\node[box] (input) {HSI patch\\$\mathbf{x}\in\R^{s\times s\times B}$};
\node[box, right=of input] (base) {Grouped spectral baseplate\\$\{\mathbf{z}_g\}_{g=1}^{G}$ and spatial context};
\node[box, right=of base] (enc) {Density-matrix encoder\\$\rho_g^{(0)}\succeq0,\ \Tr(\rho_g^{(0)})=1$};

\node[smallbox, right=8mm of enc, yshift=9mm] (spec1) {Spectral\\transition};
\node[smallbox, right=of spec1] (spa) {Spatial\\transition};
\node[smallbox, right=of spa] (coup) {Coupling\\transition};
\node[smallbox, right=of coup] (spec2) {Spectral\\transition};
\node[group, fit=(spec1)(spa)(coup)(spec2), label={[font=\footnotesize]above:Composable state-transition stack}] (stack) {};

\node[box, right=8mm of spec2, yshift=-9mm] (agg) {State aggregation\\$\rhoavg=G^{-1}\sum_g\rho_g^{(L)}$};
\node[box, right=of agg] (proto) {Prototype-state head\\$F(\rhoavg,\pi_c)$};
\node[box, right=of proto] (out) {Class prediction\\and state diagnostics};

\draw[arrow] (input) -- (base);
\draw[arrow] (base) -- (enc);
\draw[arrow] (enc.east) -- ++(4mm,0) |- (spec1.west);
\draw[arrow] (spec1) -- (spa);
\draw[arrow] (spa) -- (coup);
\draw[arrow] (coup) -- (spec2);
\draw[arrow] (spec2.east) -- ++(4mm,0) |- (agg.west);
\draw[arrow] (agg) -- (proto);
\draw[arrow] (proto) -- (out);
\end{tikzpicture}%
}
\caption{Overview of \method. Grouped spectral responses are mapped to valid density-matrix states, updated by composable spectral, spatial, and coupling transitions, and classified by fidelity to learned class-prototype states. Every transition is followed by projection onto the PSD, Hermitian, unit-trace state set.}
\label{fig:framework}
\end{figure*}

The proposed framework, termed \method, constructs one density-matrix state for each spectral group and preserves these states throughout feature refinement and classification. The states evolve through three explicit transformations: an intra-group spectral transition, a spatially conditioned transition, and an inter-group coupling transition. Each transformation is followed by a spectral projection that restores Hermiticity, PSD structure, and unit trace. The final group states are averaged and compared with learnable class-prototype density matrices using Uhlmann fidelity. Figure~\ref{fig:framework} summarizes the complete pipeline.

The contributions are as follows.
\begin{itemize}
    \item We formulate HSI classification as persistent density-matrix state learning rather than temporary matrix pooling. The learned state remains valid from encoding through classification and provides eigenspectrum-, entropy-, and purity-based diagnostics.
    \item We introduce a composable spectral-spatial-coupling transition stack that models intra-group spectral variation, local spatial context, and adjacent-group dependency in the same constrained matrix space.
    \item We replace vector-to-softmax discrimination with prototype-state matching based on density-matrix fidelity. The class representation is therefore a valid matrix state rather than an unconstrained weight vector.
    \item We report ten-run evaluation on Indian Pines, best-of-ten evaluation on WHU-Hi-LongKou, per-class analysis, classification maps, and a representation-evolution visualization under a reproducible software and hardware configuration.
\end{itemize}

\section{Related Work}

\subsection{Spectral-Spatial HSI Classification}
Early deep HSI methods learned spectral representations through stacked autoencoders and one-dimensional convolution~\cite{chen2014deep}. Joint spectral-spatial CNNs then exploited image patches and three-dimensional kernels~\cite{chen2016cnn}. SSRN introduced residual spectral and spatial blocks~\cite{zhong2018ssrn}, while HybridSN combined 3-D and 2-D convolutions~\cite{roy2020hybridsn}. SpectralFormer used transformer layers to model spectral sequence relations~\cite{hong2022spectralformer}, and recent Mamba-based models introduced selective state-space mechanisms for long-range spectral-spatial interaction~\cite{gu2023mamba,li2024mambahsi}. These methods mainly advance the feature extractor. \method\ instead focuses on the mathematical form of the persistent representation and the geometry of the classification head.

\subsection{Structured Matrix Representations}
Covariance and SPD representations encode second-order relations in matrix-valued features. SPDNet constructs a deep architecture on the SPD manifold~\cite{huang2017spdnet}; later work investigated Riemannian matrix summarization and optimization~\cite{liu2019boms,huang2021spdopt}. Density matrices share PSD structure with covariance matrices but additionally impose unit trace. The trace constraint removes arbitrary scale from the spectrum and yields bounded state diagnostics. \method\ uses a Euclidean spectral projection after every transition rather than a full Riemannian optimization routine. This choice keeps the implementation lightweight for small state dimensions while retaining explicit state validity.

\subsection{Quantum-Inspired and Prototype Learning}
Quantum-inspired models use amplitudes, phases, measurements, or density matrices as classical computational structures. Quantum-like density-matrix representations have been explored in language modeling~\cite{zhang2018nnqlm}. In HSI classification, QSSPN introduces phase prediction and measurement-like spectral-spatial fusion~\cite{zhang2023qsspn}. The distinction is not the mere presence of a density matrix. QSSPN uses it inside an intermediate fusion mechanism and then returns to vector classification; \method\ preserves matrix states through the final prediction layer. Prototype and metric learning classify samples by their proximity to class representatives~\cite{kulis2013metric,vinyals2016matching,snell2017proto}. We extend this principle from Euclidean vectors to learned density-matrix prototypes.

\begin{table}[t]
\centering
\footnotesize
\setlength{\tabcolsep}{3.5pt}
\begin{tabular}{p{0.23\columnwidth}p{0.32\columnwidth}p{0.35\columnwidth}}
\toprule
Aspect & QSSPN & \method \\
\midrule
Matrix role & Intermediate measurement-like fusion & Persistent representation from encoding to classification \\
Spectral organization & Whole input cuboid & Grouped spectral states with explicit coupling \\
State update & Pyramid feature operations & Spectral, spatial, and coupling transitions with projection \\
Classifier & Vector softmax head & Density-matrix prototypes with fidelity \\
Diagnostics & Phase/fusion response & Eigenspectrum, entropy, purity, and prototype fidelity \\
\bottomrule
\end{tabular}
\caption{Conceptual distinction between QSSPN and \method.}
\label{tab:qsspn}
\end{table}

\section{Methodology}

\subsection{Problem Formulation}
Let $\mathbf{x}_i\in\R^{s\times s\times B}$ be an HSI patch with spatial size $s\times s$ and $B$ spectral bands, and let $y_i\in\{1,\ldots,C\}$ denote its class label. Instead of learning only a Euclidean feature vector, \method\ learns a matrix state
\begin{equation}
    \mathcal{F}_{\theta}:\mathbf{x}_i\longmapsto\rhoavg_i\in\mathcal{D}(\mathcal{H}),
\end{equation}
where
\begin{equation}
\mathcal{D}(\mathcal{H})=\left\{\rho\in\C^{d_s\times d_s}:\rho\succeq0,\ \Tr(\rho)=1,\ \rho=\rho^{\dagger}\right\}.
\end{equation}
The output is compared with class prototypes $\{\pi_c\}_{c=1}^{C}\subset\mathcal{D}(\mathcal{H})$.

\subsection{Grouped Spectral Baseplate}
The $B$ bands are partitioned into $G$ contiguous groups. For group $g$, a lightweight encoder produces
\begin{equation}
    \mathbf{z}_g=f_g\!\left(\mathbf{x}[:,:,\mathcal{I}_g]\right)\in\R^{d_e},
\end{equation}
where $\mathcal{I}_g$ is the band index set of the group. In parallel, a spatial branch maps the full patch to a global context vector $\mathbf{h}_{\mathrm{ctx}}\in\R^{d_e}$. Uniform grouping is used in the reported implementation because it is deterministic and sensor-independent. The grouping operation does not assume that its boundaries exactly coincide with physical wavelength regions; its purpose is to keep local spectral continuity while making inter-group interaction explicit.

\subsection{Density-Matrix State Encoding}
A complex-valued projection reshapes each group embedding into $\mathbf{U}_g\in\C^{d_s\times d_s}$. The initial state is
\begin{equation}
\rho_g^{(0)}=
\frac{\mathbf{U}_g\mathbf{U}_g^{\dagger}+\epsilon\mathbf{I}}
{\Tr\!\left(\mathbf{U}_g\mathbf{U}_g^{\dagger}+\epsilon\mathbf{I}\right)}.
\label{eq:encoding}
\end{equation}
For any $\mathbf{v}$,
\begin{equation}
\mathbf{v}^{\dagger}\mathbf{U}_g\mathbf{U}_g^{\dagger}\mathbf{v}
=\left\|\mathbf{U}_g^{\dagger}\mathbf{v}\right\|_2^2\ge0,
\end{equation}
so the numerator is PSD; normalization gives unit trace, and the product is Hermitian. The small $\epsilon\mathbf{I}$ term prevents exactly singular states.

Let $\lambda_j(\rho)$ denote the eigenvalues of a state. Since $\lambda_j\ge0$ and $\sum_j\lambda_j=1$, the eigenspectrum behaves as a normalized mixture profile. Two state diagnostics are
\begin{equation}
\operatorname{Pur}(\rho)=\Tr(\rho^2),\qquad
S(\rho)=-\Tr(\rho\log\rho).
\end{equation}
Purity approaches one for a rank-one state and decreases as the spectrum becomes more distributed. Entropy behaves oppositely.

\subsection{Projection onto the State Set}
The transition blocks may introduce small violations of state validity. We therefore apply
\begin{equation}
\proj(\mathbf{A})=
\frac{\mathbf{Q}\,\max(\boldsymbol{\Lambda},\epsilon\mathbf{I})\,\mathbf{Q}^{\dagger}}
{\Tr\!\left[\mathbf{Q}\,\max(\boldsymbol{\Lambda},\epsilon\mathbf{I})\,\mathbf{Q}^{\dagger}\right]},
\label{eq:projection}
\end{equation}
where $(\mathbf{A}+\mathbf{A}^{\dagger})/2=\mathbf{Q}\boldsymbol{\Lambda}\mathbf{Q}^{\dagger}$. Equation~\eqref{eq:projection} symmetrizes the input, clips eigenvalues, and restores unit trace. This is a Euclidean spectral projection, not a claim that each neural block is a physical quantum channel.

\subsection{Composable State Transitions}
Let $\boldsymbol{\rho}^{(\ell)}=\{\rho_g^{(\ell)}\}_{g=1}^{G}$. The transition stack applies
\begin{equation}
\boldsymbol{\rho}^{(\ell+1)}=\mathcal{E}^{(\ell)}\!\left(\boldsymbol{\rho}^{(\ell)},\mathbf{h}_{\mathrm{ctx}}\right).
\end{equation}
The default order is spectral--spatial--coupling--spectral.

\subsubsection{Spectral Transition}
Each group is updated independently:
\begin{equation}
\rho_g'=
\proj\!\left(\mathbf{W}_{\mathrm{spec}}\rho_g\mathbf{W}_{\mathrm{spec}}^{\dagger}+\mathbf{B}_{\mathrm{spec}}\right),
\label{eq:spectral}
\end{equation}
where $\mathbf{B}_{\mathrm{spec}}$ is Hermitian. This operation changes the internal mode correlations of each spectral group.

\subsubsection{Spatial Transition}
A context modulation network produces a Hermitian matrix $\mathbf{M}_g$ from the spatial descriptor and the current state:
\begin{equation}
\rho_g'=\proj\!\left(\rho_g+\alpha_g\mathbf{M}_g(\mathbf{h}_{\mathrm{ctx}},\rho_g)\right),
\label{eq:spatial}
\end{equation}
where $\alpha_g$ is a learned gate. This block conditions spectral states on local spatial structure without flattening them.

\subsubsection{Coupling Transition}
Adjacent groups are combined through a Kronecker product:
\begin{equation}
\widetilde{\rho}_{g,g+1}=
\proj\!\left(\mathbf{V}_g(\rho_g\otimes\rho_{g+1})\mathbf{V}_g^{\dagger}\right).
\label{eq:coupling}
\end{equation}
A learned Hermitian reduction maps the coupled representation back to two $d_s\times d_s$ states, after which Eq.~\eqref{eq:projection} is applied again. The reduction is a flexible classical operation rather than a strictly completely positive trace-preserving map. Its role is to let transitions that span group boundaries influence both neighboring states.

\subsection{Prototype-State Discrimination}
The final group states are averaged:
\begin{equation}
\rhoavg=\frac{1}{G}\sum_{g=1}^{G}\rho_g^{(L)}.
\end{equation}
Each class is represented by a learnable valid state $\pi_c$. Similarity is measured by Uhlmann fidelity,
\begin{equation}
F(\rho,\sigma)=
\left[\Tr\sqrt{\sqrt{\rho}\,\sigma\,\sqrt{\rho}}\right]^2,
\label{eq:fidelity}
\end{equation}
and logits are formed as $\ell_c=F(\rhoavg,\pi_c)/\tau$, where $\tau$ is a temperature. The class probability is
\begin{equation}
p(y=c\mid\rhoavg)=
\frac{\exp(\ell_c)}{\sum_{j=1}^{C}\exp(\ell_j)}.
\label{eq:prob}
\end{equation}
The model is optimized with cross-entropy. The prototype head retains a direct semantic interpretation: prediction asks which class state has the largest fidelity with the sample state.

\subsection{Computational Cost}
State encoding requires $O(Gd_ed_s^2)$ operations. A transition stack with $L$ blocks has cost $O(LGd_s^4)$ under the small dense-matrix implementation, while fidelity computation costs $O(CGd_s^3)$. With $G=4$ and $d_s=4$, these matrix operations remain small relative to the patch encoder.

\section{Experimental Setup}

\subsection{Datasets}
We use Indian Pines and WHU-Hi-LongKou. Indian Pines was acquired by AVIRIS over an agricultural area in Indiana and contains $145\times145$ pixels, 200 commonly used spectral bands after noisy-band removal, and 16 land-cover classes~\cite{baumgardner2015indianpines}. WHU-Hi-LongKou is a UAV-borne scene with 270 spectral bands and nine crop and land-cover classes~\cite{hu2020whuhi}. Background pixels are excluded from metric computation.

\begin{table}[t]
\centering
\footnotesize
\begin{tabular}{lccc}
\toprule
Dataset & Spatial size & Bands & Classes \\
\midrule
Indian Pines & $145\times145$ & 200 & 16 \\
WHU-Hi-LongKou & $550\times400$ & 270 & 9 \\
\bottomrule
\end{tabular}
\caption{Datasets used in this preprint. Indian Pines is summarized over ten runs; WHU-Hi-LongKou reports the best of ten runs.}
\label{tab:datasets}
\end{table}

\subsection{Metrics}
Overall accuracy (OA), average accuracy (AA), Cohen's kappa, and Macro-F1 are reported. OA and AA are
\begin{equation}
\mathrm{OA}=\frac{\sum_c n_c^{\mathrm{correct}}}{\sum_c n_c},\qquad
\mathrm{AA}=\frac{1}{C}\sum_{c=1}^{C}\frac{n_c^{\mathrm{correct}}}{n_c}.
\end{equation}
AA and Macro-F1 give additional weight to minority classes and are therefore important for the imbalanced HSI setting.

\subsection{Implementation Details}
All experiments use $15\times15$ patches, $G=4$ spectral groups, group embedding dimension $d_e=16$, state dimension $d_s=4$, and the transition order Spec--Spa--Coup--Spec. The model is trained with AdamW, learning rate $10^{-3}$, weight decay $10^{-4}$, batch size 64, a maximum of 150 epochs, and early-stopping patience 20. No data augmentation or learning-rate scheduler is used in the reported runs. The fidelity temperature is $\tau=0.1$ and $\epsilon=10^{-6}$.

Experiments were executed on an NVIDIA RTX 4090 GPU and an Intel Xeon Gold 6530 CPU under Ubuntu 22.04.3. The software environment consisted of Python 3.12.12, PyTorch 2.3.1+cu121, and CUDA 12.1. The complete implementation, preprocessing scripts, split files, training commands, and evaluation code will be released publicly on GitHub with the preprint.

\section{Results and Analysis}

\subsection{Overall Performance}
Table~\ref{tab:overall} summarizes the complete-model performance. Indian Pines is reported as the mean and population standard deviation over ten runs. Following the adopted WHU-Hi-LongKou reporting protocol, the best result among ten runs is reported for that dataset.

\begin{table*}[t]
\centering
\scriptsize
\setlength{\tabcolsep}{3.2pt}
\begin{tabular}{lccccccc}
\toprule
Dataset & Reporting & OA (\%) & AA (\%) & Kappa (\%) & Macro-F1 (\%) & Best epoch & Train time (s) \\
\midrule
Indian Pines & 10-run mean$\pm$std. & $96.20\pm0.70$ & $95.57\pm1.29$ & $95.66\pm0.80$ & $93.15\pm1.11$ & $89.7\pm21.5$ & $846.7\pm159.1$ \\
WHU-Hi-LongKou & Best of 10 runs & $97.52$ & $92.49$ & $96.74$ & $92.88$ & $136$ & $858.6$ \\
\bottomrule
\end{tabular}
\caption{Classification performance of \method. Indian Pines reports all ten runs, whereas WHU-Hi-LongKou reports the best run among ten executions.}
\label{tab:overall}
\end{table*}

The Indian Pines results show stable performance across random seeds: OA varies by less than one percentage point in standard deviation, while AA remains above $95\%$ on average. The difference between OA and Macro-F1 reflects the strong class imbalance of the scene. On WHU-Hi-LongKou, the best run reaches $97.52\%$ OA and $92.49\%$ AA. The lower AA relative to OA again indicates that the smallest classes remain more difficult than the dominant land-cover categories.

\subsection{Run-to-Run Stability}
Table~\ref{tab:ip_runs} lists all ten Indian Pines runs. No single seed dominates every metric: run 8 gives the highest OA, run 3 gives the highest AA, and run 4 gives the highest Macro-F1. This metric-dependent ordering is a consequence of class imbalance and shows why the ten-run aggregate is more informative than selecting one operating point. The best epoch ranges from 59 to 134, but the final OA remains within a narrow interval of approximately $2.1$ percentage points.

\begin{table*}[t]
\centering
\scriptsize
\setlength{\tabcolsep}{5pt}
\begin{tabular}{crrrrrr}
\toprule
Run & Seed & OA (\%) & AA (\%) & Kappa (\%) & Macro-F1 (\%) & Best epoch \\
\midrule
1 & 42 & 95.10 & 94.62 & 94.41 & 91.79 & 69 \\
2 & 142 & 96.63 & 94.99 & 96.16 & 94.55 & 94 \\
3 & 242 & 96.98 & 97.71 & 96.55 & 93.24 & 90 \\
4 & 342 & 95.98 & 96.45 & 95.41 & 95.31 & 91 \\
5 & 442 & 95.84 & 96.26 & 95.26 & 93.18 & 116 \\
6 & 542 & 97.10 & 95.46 & 96.69 & 93.36 & 134 \\
7 & 642 & 95.74 & 96.01 & 95.16 & 91.46 & 93 \\
8 & 742 & 97.18 & 92.69 & 96.79 & 92.89 & 85 \\
9 & 842 & 96.05 & 96.45 & 95.50 & 93.46 & 59 \\
10 & 942 & 95.37 & 95.07 & 94.72 & 92.27 & 66 \\
\midrule
Mean$\pm$std. & -- & $96.20\pm0.70$ & $95.57\pm1.29$ & $95.66\pm0.80$ & $93.15\pm1.11$ & $89.7\pm21.5$ \\
\bottomrule
\end{tabular}
\caption{Indian Pines results for all ten random seeds.}
\label{tab:ip_runs}
\end{table*}

\subsection{Per-Class Performance}
Table~\ref{tab:perclass} reports the mean and standard deviation of Indian Pines class accuracies across ten runs, together with the class accuracies of the best WHU-Hi-LongKou run. The rare Indian Pines classes, especially Oats and Grass-pasture-mowed, exhibit larger variance. This pattern is expected because a small change in the number of correct predictions causes a large percentage change for classes with few labeled samples. On WHU-Hi-LongKou, Narrow-leaf soybean, Sesame, and Mixed weed are the most difficult classes in the selected best run.

\begin{table*}[t]
\centering
\scriptsize
\setlength{\tabcolsep}{3.4pt}
\begin{tabular}{rlr@{\hspace{7mm}}rlr}
\toprule
\multicolumn{3}{c}{Indian Pines: 10-run class accuracy (\%)} & \multicolumn{3}{c}{WHU-Hi-LongKou: best-run class accuracy (\%)} \\
\cmidrule(lr){1-3}\cmidrule(lr){4-6}
ID & Class & Mean$\pm$std. & ID & Class & Accuracy \\
\midrule
1 & Alfalfa & $97.78\pm3.24$ & 1 & Corn & 99.38 \\
2 & Corn-notill & $95.23\pm1.56$ & 2 & Cotton & 97.06 \\
3 & Corn-mintill & $95.44\pm1.97$ & 3 & Sesame & 86.05 \\
4 & Corn & $94.23\pm4.56$ & 4 & Broad-leaf soybean & 97.78 \\
5 & Grass-pasture & $94.42\pm1.65$ & 5 & Narrow-leaf soybean & 80.93 \\
6 & Grass-trees & $98.85\pm1.10$ & 6 & Rice & 97.06 \\
7 & Grass-pasture-mowed & $93.18\pm9.37$ & 7 & Water & 99.89 \\
8 & Hay-windrowed & $99.45\pm0.52$ & 8 & Roads and houses & 87.64 \\
9 & Oats & $88.75\pm14.74$ & 9 & Mixed weed & 86.66 \\
10 & Soybean-notill & $93.55\pm1.89$ & & & \\
11 & Soybean-mintill & $96.90\pm1.18$ & & & \\
12 & Soybean-clean & $94.78\pm2.87$ & & & \\
13 & Wheat & $99.21\pm0.39$ & & & \\
14 & Woods & $98.49\pm0.86$ & & & \\
15 & Buildings--Grass--Trees--Drives & $91.04\pm4.01$ & & & \\
16 & Stone--Steel--Towers & $97.87\pm5.10$ & & & \\
\bottomrule
\end{tabular}
\caption{Per-class accuracy on Indian Pines and WHU-Hi-LongKou. Indian Pines values are computed across ten runs; WHU-Hi-LongKou values are from the selected best run.}
\label{tab:perclass}
\end{table*}

\subsection{Classification Maps and Representation Evolution}
\begin{figure*}[t]
\centering
\subfloat[Reference map]{\includegraphics[width=0.155\textwidth]{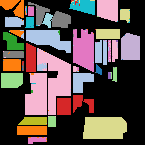}}
\hfill
\subfloat[Prediction map]{\includegraphics[width=0.155\textwidth]{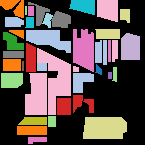}}
\hfill
\subfloat[Representation before and after the state-transition stack]{\includegraphics[width=0.63\textwidth]{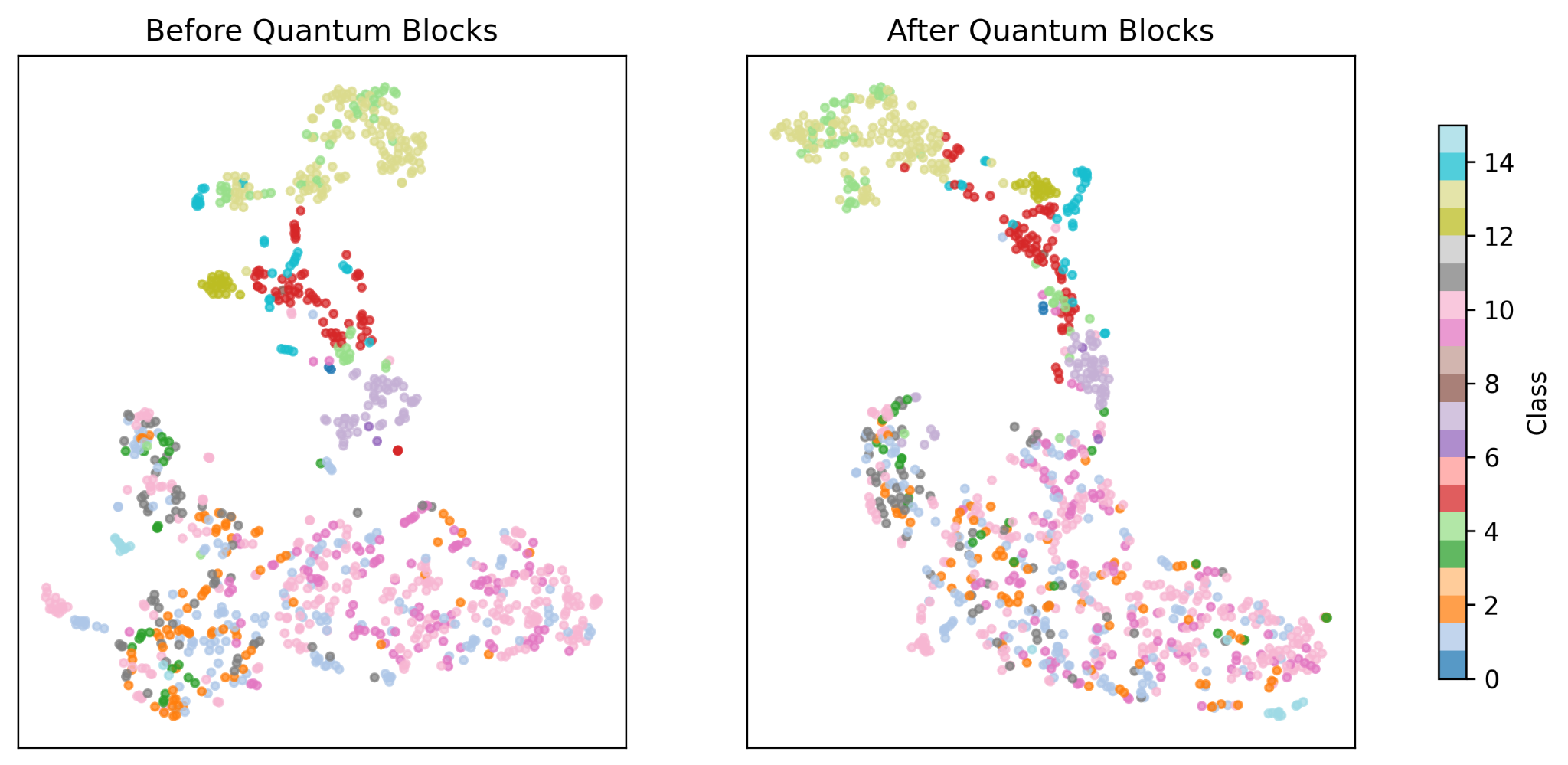}}
\caption{Qualitative results on Indian Pines. The first two panels compare ground truth and prediction. The third panel shows a two-dimensional projection of the learned representation before and after the transition blocks.}
\label{fig:qualitative}
\end{figure*}

Figure~\ref{fig:qualitative} compares the Indian Pines reference map with the predicted map. The major agricultural and woodland regions are recovered with coherent spatial structure. Most visible errors occur at thin boundaries and within small classes, consistent with the per-class variance in Table~\ref{tab:perclass}.

The third panel visualizes the feature distribution before and after the transition stack. The projection is used only for visualization and does not enter training. Before the transitions, several classes overlap in the two-dimensional embedding. After the spectral, spatial, and coupling operations, the distribution contains more compact class regions and clearer local separation. This observation is consistent with the intended role of the transition stack: to reorganize grouped matrix states before prototype matching rather than simply add a final vector classifier.

\subsection{Interpretation of the Matrix State}
The model's diagnostics arise from the state constraints rather than from an additional post-hoc explainer. The eigenvalues of $\rhoavg$ sum to one, so changes in the spectrum can be compared between samples without a separate norm calibration. Entropy increases when the spectrum is distributed over several modes, whereas purity increases when one mode dominates. Prototype fidelity measures how closely a sample state matches each class state. These quantities do not by themselves prove physical material abundance, and we do not interpret them as a substitute for hyperspectral unmixing. Their value is that they provide internally consistent, bounded readouts of the learned representation.

\subsection{Discussion and Limitations}
The results establish the feasibility of persistent density-matrix state learning for HSI classification, but several limitations should be made explicit. First, the current experiments use a fixed state dimension $d_s=4$ and uniform spectral grouping. Larger states and adaptive group boundaries may improve capacity at additional computational cost. Second, the learned coupling reduction is flexible but is not constrained to be a physical quantum channel. Third, the present preprint reports complete-model results and a qualitative representation-evolution analysis; a controlled component-wise ablation and boundary-pixel entropy study would provide stronger causal evidence for individual design choices. Finally, the two datasets are agricultural scenes. Evaluation on urban and cross-sensor data would better test transferability.

These limitations do not affect the main distinction from existing vector-based classifiers: the representation and the class prototypes remain valid density matrices throughout the network. They instead define the next experiments needed to determine which part of the structured state space is most responsible for the observed performance.

\section{Conclusion}
This paper introduced \method, a density-matrix representation learning framework for hyperspectral image classification. Grouped spectral responses are encoded as PSD, Hermitian, trace-normalized states; spectral, spatial, and coupling transitions refine these states; and prediction is performed by fidelity to learned class-prototype density matrices. The formulation provides normalized spectral diagnostics while remaining fully executable on classical hardware. Ten-run experiments on Indian Pines produce $96.20\pm0.70\%$ OA, and the best of ten WHU-Hi-LongKou runs reaches $97.52\%$ OA. Per-class analysis, classification maps, and projected representations show both the strengths of the method and the remaining difficulty of minority and boundary classes. The study indicates that constrained matrix-state learning is a viable direction for multidimensional image recognition and motivates broader evaluation of state dimension, grouping, and mixed-pixel diagnostics.

\section*{Code Availability}
The complete source code, preprocessing and split scripts, training commands, evaluation routines, and figure-generation utilities will be released publicly on GitHub with the preprint. The repository link is omitted from this working copy.

\bibliographystyle{IEEEtran}
\bibliography{references}

\end{document}